\documentclass[11pt]{article}

\usepackage[final]{acl}

\usepackage{times}
\usepackage{latexsym}
\usepackage[T1]{fontenc}
\usepackage[utf8]{inputenc}
\usepackage{microtype}
\usepackage{inconsolata}
\usepackage{graphicx}
\usepackage{booktabs}
\usepackage{amsmath}
\usepackage{amssymb}
\usepackage{algorithm}
\usepackage{algpseudocode}
\usepackage{multirow}
\usepackage{xcolor}
\usepackage{colortbl}
\usepackage{xspace}
\usepackage{enumitem}

\newcommand{\method}{\textsc{Skin-Deep}\xspace}
\newcommand{\gfs}{\textsc{GFS}\xspace}
\definecolor{groupbg}{HTML}{FFF2CC}
\definecolor{aclblue}{HTML}{000099}
\hypersetup{colorlinks=true,citecolor=aclblue,linkcolor=aclblue,urlcolor=aclblue}
\newcommand{\grouprow}[2]{\rowcolor{groupbg}[0pt][0pt]\multicolumn{#1}{@{}c@{}}{\textit{#2}\strut}}
\definecolor{refusefg}{HTML}{1F6B45}
\definecolor{refusebg}{HTML}{E9F4EC}
\definecolor{complyfg}{HTML}{A3312F}
\definecolor{complybg}{HTML}{FBECEA}
\definecolor{withheldfg}{HTML}{7A7F87}
\newcommand{\refuse}[1]{\textcolor{refusefg}{\textbf{#1}}}
\newcommand{\comply}[1]{\textcolor{complyfg}{\textbf{#1}}}
\newcommand{\withheld}{\mbox{\textcolor{withheldfg}{[\,$\cdots$ withheld\,]}}}
\newlength{\qualw}
\newcommand{\promptcat}[1]{\textbf{#1}}

\title{\textsc{Skin-Deep}: A Geometric Diagnostic for Alignment Fragility in Large Language Model Representations}

\author{
  \bfseries Dongyub Jude Lee$^{1}$\thanks{Equal contribution.} \quad
  Jungseob Lee$^{2}$\footnotemark[1] \quad
  Seungyoon Lee$^{2}$ \quad
  Seongtae Hong$^{2}$ \\
  \bfseries Suhyune Son$^{2}$ \quad
  Sugyeong Eo$^{3}$ \quad
  Jaehyung Seo$^{4}$\thanks{Corresponding authors.} \quad
  Heuiseok Lim$^{2}$\footnotemark[2] \\
  $^{1}$Zoom Communications \quad $^{2}$Korea University \quad $^{3}$Yonsei University \quad $^{4}$Konkuk University \\
  \texttt{jude.lee@zoom.us} \\
  \texttt{\{omanma1928, dltmddbs100, ghdchlwls123, ssh5131, limhseok\}@korea.ac.kr} \\
  \texttt{s.eo@yonsei.ac.kr, seojae777@konkuk.ac.kr}
}

\begin{document}
\maketitle

\begin{abstract}
Refusal on a safety benchmark does not reveal how stable that behavior will remain after model updates. Benign downstream fine-tuning can weaken refusal, yet behavioral evaluations typically expose this fragility only after an intervention. We introduce \method, a geometric diagnostic that examines the unmodified model's residual-stream activations. It compares aligned and base checkpoints to identify safety-separating directions, tests their behavioral relevance through ablation, and summarizes the layer-wise pattern in the Geometric Fragility Score (\gfs). Across twenty-one instruction-tuned models, harmful requests and benign instructions exhibit a recurring low-rank separation pattern. Selected direction ablations weaken refusal, with the effective direction varying across models. In benign low-rank fine-tuning experiments, the initially safe model with the lowest score before fine-tuning has the lowest harmful-compliance rate when trained on the largest tested set of harmless examples. These findings connect representation geometry to subsequent behavioral susceptibility and support activation-based diagnostics as a complement to refusal tests. Our code is available at \url{https://github.com/js-lee-AI/skin-deep}.
\end{abstract}

\section{Introduction}

Post-training alignment is the main route from pretrained language models to deployed assistants \citep{ouyang2022training,bai2022training,lee2022norose}. Behavioral safety tests assess whether an assistant refuses harmful requests \citep{mazeika2024harmbench,lee2026agentloop}. However, a refusal on a tested prompt does not establish that the underlying behavior will survive changes to the prompt or model weights. Studies of inference-time attacks and downstream fine-tuning show that apparently safe checkpoints can become compliant with harmful requests after small interventions \citep{zou2023universal,wei2023jailbroken,qi2024finetuning,yang2023shadow,lermen2024lora}. We call this susceptibility \emph{alignment fragility}.

\begin{figure*}[t]
\centering
\includegraphics[width=\textwidth]{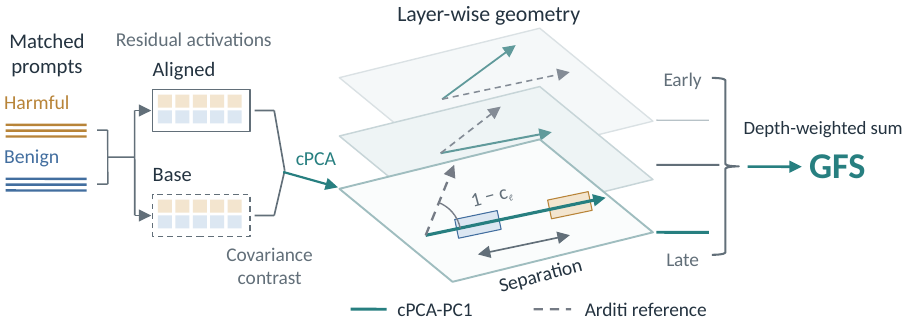}
\caption{\method contrasts aligned and base covariances on matched harmful and benign prompts. Projected separation, angular departure, and depth contribute to \gfs before downstream fine-tuning. Here $c_\ell$ is the absolute cPCA-PC1--Arditi cosine. Layer geometry is schematic.}
\label{fig:overview}
\end{figure*}

Figure~\ref{fig:overview} sketches our representation-level approach to assessing fragility. A behavioral test observes the current response, whereas a fragility test usually requires applying the intervention whose consequences the assessor wants to anticipate. A representation-level diagnostic could complement these tests by examining how refusal is encoded before changing the model. Evidence that refusal depends on low-dimensional residual-stream directions motivates this possibility \citep{arditi2024refusal,panickssery2023steering}. The remaining question is how to connect such geometry to behavioral susceptibility across models without equating prompt-category separation with a refusal mechanism.

We introduce \method, a diagnostic that compares aligned and base-model activations on matched harmful requests and benign instructions. Contrastive principal component analysis identifies directions whose activation variance is emphasized by alignment. We then measure harmful/benign separation along those directions, test selected directions through ablation, and compare activation relationships across model families. Finally, the Geometric Fragility Score (\gfs) summarizes the separation, its depth, and its relation to a known refusal direction. Computing this score requires activation extraction \citep{lee2026bytecross} but no adversarial optimization or downstream fine-tuning.

Our study covers twenty-one instruction-tuned models, using overlapping subsets for each analysis. It supports four connected findings.
\begin{itemize}[leftmargin=*]
\item \textbf{Separation.} Harmful and benign prompts separate along a small set of directions. In split-sample checks, directions fitted on one prompt subset also separate held-out prompts.
\item \textbf{Behavioral relevance.} Removing selected PCA, contrastive PCA, or reference refusal directions reduces harmful-request refusal in several models. The effective direction varies by model.
\item \textbf{Recurrence.} Different model families preserve similar activation relationships among matched prompts. Their strongest separation occurs at different depths, so recurrence does not require a shared layer schedule.
\item \textbf{Diagnostic association.} Within the initially safe models tested with benign low-rank fine-tuning, the lowest-scoring model has the lowest harmful-compliance rate after training on the largest tested fine-tuning set. This connects a score computed before fine-tuning to a subsequent behavioral distinction.
\end{itemize}

Together, these findings link activation geometry to refusal behavior and show how representation measurements can complement behavioral assessment before downstream fine-tuning.

\section{Related Work}
\label{sec:related}

\paragraph{Behavioral fragility.} Adversarial suffixes \citep{zou2023universal} and role-play jailbreaks \citep{wei2023jailbroken} bypass refusal without changing model weights. Weight-level interventions can weaken refusal through downstream fine-tuning \citep{qi2024finetuning,yang2023shadow}, low-rank adapters trained on tens of examples \citep{lermen2024lora}, or output-layer pruning \citep{wei2024assessing}. These studies establish behavioral susceptibility by applying an intervention. Our question is whether activations from the unmodified checkpoint carry a useful diagnostic signal for this susceptibility.

\paragraph{Refusal representations.} \citet{arditi2024refusal} identifies a residual-stream direction from the difference between mean harmful and harmless activations and shows that ablating it can disable refusal. We call this the \emph{Arditi direction}. Representation engineering \citep{zou2023representation} and contrastive activation addition \citep{panickssery2023steering} also use activation directions to steer model behavior. \citet{lee2024mechanistic} studies how Direct Preference Optimization (DPO) \citep{arxiv-2305-18290} reduces the expression of toxic features without erasing them. These findings motivate testing whether refusal is associated with a broader low-rank geometry and whether a summary of that geometry distinguishes models before fine-tuning.

\paragraph{Activation measurements.} Linear probes \citep{alain2017understanding,lee2026layermix} measure what information is available at individual layers. Contrastive PCA \citep{abid2018exploring} isolates directions emphasized in one activation population relative to another. Centered kernel alignment (CKA) compares activation relationships across networks \citep{kornblith2019similarity,ding2021grounding}. \method combines these measurements with direction ablations and a downstream fine-tuning test.

\section{Geometric Fragility}
\label{sec:geometricfragility}

\subsection{Diagnostic Setting}
\label{sec:preliminaries}

We call refusal \emph{geometrically fragile} when it is associated with a low-rank activation pattern that selected direction interventions can disrupt. \method first searches for directions that separate harmful requests from benign instructions, then tests their behavioral relevance. The resulting score summarizes the unmodified model, and the intervention experiments connect that geometry to changes in refusal.

Let $M$ be an aligned causal transformer and $M_0$ its version-matched base counterpart, with $L$ layers and hidden size $d$. For input $x$, let $h_\ell(x),h^0_\ell(x)\in\mathbb{R}^d$ be their residual-stream activations at layer $\ell$. The safety set $\mathcal{D}_{\mathrm{safe}}$ contains harmful requests that an aligned assistant should refuse. The general set $\mathcal{D}_{\mathrm{gen}}$ contains benign instructions. These names describe the intended response to a prompt, not an observed refusal label.

At each layer, we collect activations at the final token for matched prompt sets and center the resulting matrices $H^{\mathrm{inst}}_\ell$ and $H^{\mathrm{base}}_\ell$. Their covariance matrices are $\Sigma^{\mathrm{inst}}_\ell$ and $\Sigma^{\mathrm{base}}_\ell$. Matching the prompts across aligned and base checkpoints keeps the comparison tied to the same inputs. Matching harmful and benign prompt lengths reduces a separate source of activation variation.

\subsection{Directions and Behavioral Tests}
\label{sec:method}

\paragraph{Subspace search.} Contrastive PCA (cPCA) \citep{abid2018exploring} finds a unit direction with large aligned-model variance relative to base-model variance,
\begin{equation}
\label{eq:cpca}
\mathbf{v}^{\mathrm{cPCA}}_\ell=\arg\max_{\|\mathbf{v}\|=1}\mathbf{v}^{\top}
\left(\Sigma^{\mathrm{inst}}_\ell-\alpha\Sigma^{\mathrm{base}}_\ell\right)\mathbf{v}.
\end{equation}
The contrastive covariance emphasizes directions whose variance is larger in the aligned model relative to the base model. Both covariance matrices use the same prompt mixture, so their difference isolates the contrast between checkpoints. The parameter $\alpha$ determines how strongly base-model variance is penalized. The leading eigenvector supplies the cPCA first principal component (cPCA-PC1), while the leading eigenvectors together define the subspace used for the overlap analysis.

For a fitted direction, let $z_\ell(x)=\mathbf{v}_\ell^\top h_\ell(x)$ be the scalar projection. Cohen's $d$ \citep{cohen2013statistical} measures the difference between harmful and benign projected means relative to their pooled standard deviation,
\begin{equation}
\label{eq:cohend}
 d_\ell=\frac{\mu_{\mathrm{safe},\ell}-\mu_{\mathrm{gen},\ell}}{s_{\mathrm{pooled},\ell}}.
\end{equation}
The means and pooled deviation are computed from the corresponding projected groups. Taking $|d_\ell|$ makes the result invariant to the arbitrary sign of an eigenvector. This gives a common scale for comparing separation across layers without treating a classifier threshold as part of the score. The held-out check fits the direction and measures its projected separation on different prompts.

We distinguish this one-dimensional measurement from evidence in the full activation space. Permutational multivariate analysis of variance (PERMANOVA) \citep{anderson2001permanova} assesses the harmful/benign split without projection onto cPCA-PC1. PCA on unit-normalized activations checks whether separation persists after removing activation magnitude. We use these full-space and norm measurements as complementary validation of the projected signal.

For the overlap analysis, let $Q_\ell$ contain an orthonormal basis for the top-five cPCA subspace. The Arditi direction $\mathbf{v}^{\mathrm{Arditi}}_\ell$ is the harmful-minus-benign residual-stream mean difference at that layer, normalized to unit length. The smallest principal angle satisfies
\begin{equation}
\label{eq:overlap}
 \cos\theta_\ell=\|Q_\ell^\top\mathbf{v}^{\mathrm{Arditi}}_\ell\|_2.
\end{equation}
A small angle places the reference direction close to the recovered subspace. This quantity differs from the cosine with cPCA-PC1 alone. The subspace can contain a reference direction even when that direction is distributed across several components. We retain both measurements to describe the broader subspace and the single direction entering the score.

\paragraph{Direction ablation.} At a model-specific peak-separation layer $\ell^\star$, we project a unit direction out of the residual stream during generation,
\begin{equation}
\label{eq:ablation}
\widetilde h_{\ell^\star}=h_{\ell^\star}-
\left(\mathbf{v}_{\ell^\star}^{\top}h_{\ell^\star}\right)\mathbf{v}_{\ell^\star}.
\end{equation}
We compare PCA-PC1, cPCA-PC1, the Arditi direction, and a random direction under this same intervention. A reduction in refusal supports a behavioral role for the direction tested. The random direction controls for an unrelated residual-stream edit, and a non-leading PCA direction provides an additional specificity control. Comparing these interventions identifies which directions affect refusal within each model.

\paragraph{Cross-model recurrence.} Linear CKA compares the activation matrices of two models on the same prompts at their respective peak layers \citep{kornblith2019similarity}. It measures whether prompt-to-prompt activation relationships agree, even when hidden dimensions differ. A prompt-shuffle null breaks prompt correspondence before recomputing CKA. Depth-normalized separation profiles provide a distinct check of whether models also share a layer schedule. Together, these comparisons distinguish recurrence of prompt relationships from agreement about where the strongest signal appears.

\subsection{Geometric Fragility Score}
\label{sec:gfs}

\gfs summarizes the layer-wise pattern using separation, depth, and angular distance from the known refusal axis,
\begin{equation}
\label{eq:gfs}
\gfs(M)=\sum_{\ell=1}^{L}w_\ell |d_\ell|
\bigl(1-|\cos(\mathbf{v}_\ell,\mathbf{v}^{\mathrm{Arditi}}_\ell)|\bigr),
\end{equation}
where $\mathbf{v}_\ell$ is cPCA-PC1, $d_\ell$ is its projected Cohen's $d$, and the default depth weight is $w_\ell=\ell/L$. The separation term gives more weight to layers where harmful and benign prompts differ. The depth weight emphasizes later layers. The cosine term downweights cPCA directions that align with the Arditi direction, so a high score indicates separation that extends away from that reference axis.

The default aggregation sums layer contributions without normalization. Its total weight is $\sum_{\ell=1}^{L}\ell/L=(L+1)/2$, so \gfs reflects both the separation-and-overlap profile and model depth.

Algorithm~\ref{alg:gfs} summarizes the computation. Activation extraction and the geometric measurements precede any weight update or adversarial optimization. The resulting evidence table retains the effect sizes, full-space tests, and reference alignment, making the scalar traceable to its layer-level components. Scores are compared within an extraction protocol, since prompt formatting and token position change the activations being summarized.

\begin{algorithm}[t]
\small
\caption{\textsc{Skin-Deep} diagnostic computation}
\label{alg:gfs}
\begin{algorithmic}[1]
\Require Aligned model $M$, base model $M_0$, matched prompt sets $\mathcal{D}_{\rm safe},\mathcal{D}_{\rm gen}$
\Ensure Evidence table $E$ and scalar \gfs
\For{$\ell=1,\ldots,L$}
\State Extract and center $H_\ell,H^0_\ell$ on matched prompts
\State Fit $\mathbf{v}_\ell$ by cPCA using Eq.~\ref{eq:cpca}
\State Compute projected harmful/benign effect $d_\ell$
\State Compute full-space PERMANOVA $p_\ell$
\State $c_\ell\gets |\cos(\mathbf{v}_\ell,\mathbf{v}^{\rm Arditi}_\ell)|$
\EndFor
\State $p^*_{1:L}\gets\operatorname{BH\text{-}FDR}(p_{1:L},q=0.05)$
\State $\gfs\gets\sum_{\ell=1}^{L}(\ell/L)|d_\ell|(1-c_\ell)$
\State $E\gets\{(d_\ell,p^*_\ell,c_\ell)\}_{\ell=1}^{L}$
\State \Return $E$ and \gfs
\end{algorithmic}
\end{algorithm}

\section{Experimental Setup}
\label{sec:experimental-setup}

\subsection{Models}

The full pool contains twenty-one instruction-tuned models with 3B--32B parameters. The reported training labels include supervised fine-tuning (SFT) \citep{lee2025clo,lee2026answerconditioned}, reinforcement learning from human feedback combined with SFT (RLHF+SFT) \citep{ouyang2022training}, DPO, Odds Ratio Preference Optimization (ORPO) \citep{arxiv-2403-07691}, reinforcement learning from AI feedback (RLAIF) \citep{arxiv-2309-00267}, and Conditioned Reinforcement Learning Fine-Tuning (C-RLFT) \citep{arxiv-2309-11235}. We use IFT for Mistral-v0.3 and Yi-1.5-9B-Chat to denote instruction fine-tuning without specifying a preference-optimization stage. These labels characterize checkpoints, rather than controlled comparisons of training algorithms.

Five overlapping cohorts serve different analyses. The \emph{core set} contains Llama-3.1-8B, Qwen-2.5-7B, Mistral-7B-v0.3, and Gemma-2-9B \citep{dubey2024llama3,qwen2024qwen25,jiang2023mistral,gemma2024team}. It supports the principal separation measurements, cross-model CKA, and replicated LoRA analysis. The \emph{raw-prompt ranking set} contains sixteen models. The \emph{chat-template set} contains eight models spanning several training labels. The \emph{ablation set} contains six models, including all four core models, Mistral-7B-Instruct-v0.2, and SOLAR-10.7B. Finally, the \emph{LoRA set} adds Tulu-3-8B-DPO, Qwen-2.5-3B, and Qwen-2.5-14B to the core set. We define an initially safe checkpoint by pre-LoRA harmful compliance at most $0.1$. All three additions satisfy this criterion.

Analysis tables identify the models actually evaluated. A model's inclusion in one cohort does not imply that it was measured in every other protocol. Hybrid-reasoning Qwen3 models \citep{lee2026dart_routing,lee2026acoer} are excluded from the raw-prompt ranking because their extraction condition depends on the model-specific thinking flag. Appendix~\ref{sec:appendix-model-ranking} gives the scope of this exclusion.

\subsection{Prompts and Activations}

Activation measurements use 500 harmful requests and 500 benign instructions matched by token count. The harmful set combines 200 AdvBench prompts, 200 HarmBench prompts, and 100 BeaverTails prompts \citep{zou2023universal,mazeika2024harmbench,ji2023beavertails}. The benign set combines 250 Alpaca instructions and 250 OASST instructions \citep{taori2023alpaca,kopf2023openassistant}.

For the raw-prompt ranking, inputs are passed directly to the model and activations are extracted at the final token. For the chat-template analysis, model-specific chat formatting is applied first and activations are taken at the final attended token. Models enter the chat-template cohort when their baseline refusal rate judged by Llama-Guard-3-8B lies in $[0.30,0.90]$ \citep{inan2023llamaguard,dubey2024llama3,lee2026madara}. This selection criterion belongs to that extraction analysis. We retain all six ablation models, including models with low or high baseline refusal under its separate evaluation.

cPCA uses $\alpha=100$. This value is selected from $\alpha\in\{1,10,20,50,100,200,500,1000\}$ using harmful/benign probe accuracy on a held-out 20\% split. Linear probes use logistic regression with five-fold cross-validation. Token-position checks compare the final attended token, the first attended token other than the beginning-of-sequence (BOS) token, and the mean over attended tokens.

\subsection{Behavior and Uncertainty}

Direction ablation removes the selected direction at a peak-separation layer during generation. Harmful-request refusal is evaluated on 100 AdvBench prompts. The main comparison includes PCA-PC1, cPCA-PC1, Arditi, and random directions. PCA-PC4 is additionally tested for Llama.

LoRA fine-tuning uses rank-eight adapters on harmless Alpaca subsets \citep{hu2022lora}. Reported subset sizes are $n\in\{5,10,25,50,100,150,200\}$. Harmful compliance is measured on 50 held-out harmful prompts using a rule-based harm-string match together with Llama-Guard. We use three random seeds for Llama, Qwen, and Mistral, and for Gemma at $n=200$. The other Gemma entries and the three added models provide the reported trajectories without a multi-seed uncertainty estimate. Refusal and compliance rates from the two behavioral protocols are not complements because their prompt populations and judges differ.

PERMANOVA uses 1000 permutations, with Benjamini--Hochberg false discovery rate correction at $q<0.05$ within each model \citep{benjamini1995controlling}. Peak descriptive separation is selected across layers and examined components. We report it separately from split-sample cPCA estimates, which fit directions on one half of the prompts and evaluate separation on the held-out half. The split-sample separation intervals use 50 bootstrap resamples. Ablation rates have marginal Wilson 95\% confidence intervals. A marked refusal decrease requires the post-ablation upper bound to lie below the baseline lower bound. Appendix~\ref{sec:robustness-selection} details the selection check.

\section{Results}
\label{sec:results}

\subsection{Safety-Separating Geometry}
\label{sec:results-c1}

Table~\ref{tab:c1} reports held-out cPCA separation and linear-probe accuracy for the core models. In the split-sample check, directions fitted on one prompt subset separate the other subset, with bootstrap intervals above zero for all four models.

\begin{table}[t]
\centering
\small
\begin{tabular*}{\columnwidth}{@{\extracolsep{\fill}}lcc@{}}
\toprule
Model & Held-out $d$ [95\% CI] & Probe acc. \\
\midrule
Llama-3.1-8B & $3.23\;[2.96,3.51]$ & 0.937 \\
Qwen-2.5-7B & $2.97\;[2.68,3.24]$ & 0.959 \\
Mistral-7B-v0.3 & $2.71\;[2.50,2.95]$ & 0.904 \\
Gemma-2-9B & $2.88\;[2.56,3.12]$ & 0.926 \\
\bottomrule
\end{tabular*}
\caption{Held-out cPCA separation and linear-probe accuracy for the core models. Separation estimates include 95\% bootstrap percentile intervals.}
\label{tab:c1}
\label{tab:c1-ci}
\end{table}

Table~\ref{tab:c1} gives the core estimates within a broader fourteen-model split-sample check. Across that check, the mean train-minus-test gap is $+0.070$, and nine models have an absolute gap below $0.08$. The minimum held-out effect is $1.69$ for Qwen-2.5-3B. Thus the held-out separation extends beyond the four core models. Notus-7B-v1 and Mistral-7B-Instruct-v0.1 enter the raw-prompt ranking with point estimates and are outside this bootstrap summary.

The full-space tests and norm controls complement this finding. PERMANOVA remains significant after correction for all core models, so the harmful/benign distinction is visible outside a single projection. Unit-normalized PCA also retains separation, which rules out activation magnitude as its sole explanation.

Table~\ref{tab:c1-extended} shows that the separation persists under model-specific chat formatting. The signal appears across the eight-model cohort, including different training labels. The peak location also matters because an early peak can capture input-format structure rather than the representation used to generate a refusal. We therefore examine extraction position and subspace overlap alongside the size of the peak.

\begin{table*}[t]
\centering
\small
\begin{tabular*}{\textwidth}{@{\extracolsep{\fill}}llccccc@{}}
\toprule
Model & Training label & Peak $|d|$ & Peak layer & Relative depth & $\theta$ & $\cos\theta$ \\
\midrule
Mistral-7B-Instruct-v0.2 & SFT & 7.031 & 22/32 & 0.69 & $11.3^\circ$ & 0.98 \\
Mistral-7B-Instruct-v0.3 & IFT & 6.157 & 19/32 & 0.59 & $8.8^\circ$ & 0.99 \\
Yi-1.5-9B-Chat & IFT & 5.078 & 46/48 & 0.96 & $8.1^\circ$ & 0.99 \\
openchat-3.5-1210 & C-RLFT & 5.012 & 12/32 & 0.38 & $17.4^\circ$ & 0.95 \\
Hermes-2-Pro-Mistral-7B & DPO & 4.941 & 13/32 & 0.41 & $13.1^\circ$ & 0.97 \\
SOLAR-10.7B-Instruct & DPO & 4.713 & 46/48 & 0.96 & $29.1^\circ$ & 0.87 \\
Starling-LM-7B-alpha & RLAIF & 2.653 & 31/32 & 0.97 & $76.7^\circ$ & 0.23 \\
Nous-Hermes-2-Mistral-7B-DPO & DPO & 2.520 & 1/32 & 0.03 & $40.8^\circ$ & 0.76 \\
\bottomrule
\end{tabular*}
\caption{Chat-template separation and subspace overlap. $\theta$ is the smallest principal angle between the top-five cPCA subspace and the Arditi direction. Its cosine measures projected overlap with the subspace.}
\label{tab:c1-extended}
\label{tab:grassmann}
\end{table*}

Table~\ref{tab:tokenpos} shows that first-token separation is weak, whereas both final-token and mean-pooled activations retain a substantial signal. Mean pooling is useful here because it does not depend on a single final position. The result is consistent with information accumulating across the prompt, while leaving the strength and layer location sensitive to the extraction protocol. This five-model position check is separate from the eight-model chat-template summary. Appendix~\ref{sec:appendix-token-position} describes the cohort boundary.

\begin{table}[t]
\centering
\small
\setlength{\tabcolsep}{4pt}

\begin{tabular*}{\columnwidth}{@{\extracolsep{\fill}}lccc@{}}
\toprule
\textbf{Model} & \textbf{Final} & \textbf{First} & \textbf{Mean} \\
\midrule
Nous-Hermes-2-Mistral & 2.78 & 0.22 & 1.98 \\
Mistral-v0.2     & 6.61 & 0.23 & 2.78 \\
Mistral-v0.3     & 7.12 & 0.26 & 2.05 \\
Llama-3.1-8B        & 7.32 & 0.24 & 2.89 \\
Qwen-2.5-7B         & 7.35 & 0.20 & 4.63 \\
\bottomrule
\end{tabular*}
\caption{Peak cPCA separation at the final attended token, first attended non-BOS token, and mean over attended tokens after chat formatting.}
\label{tab:tokenpos}
\end{table}

Table~\ref{tab:c1-extended} also shows that the Arditi direction lies within $20^\circ$ of the top-five cPCA subspace in five of the eight models. This supports overlap between the recovered subspace and a known refusal direction, rather than equivalence between cPCA-PC1 and that direction. For Starling, substantial separation coexists with weak overlap.

\subsection{Direction-Specific Refusal Changes}
\label{sec:results-c2}

Table~\ref{tab:c2} shows a detectable refusal decrease for at least one tested direction in four of the six ablation models. The effective intervention differs by model. PCA-PC1 produces the clearest decrease for Llama, whereas the Arditi direction does so for Qwen and SOLAR. Mistral-v0.2 responds to all three non-random directions. Random-direction rates remain close to the corresponding baselines. This comparison supports direction-specific effects rather than a uniform consequence of modifying the residual stream.

\begin{table}[t]
\centering
\fontsize{8}{9.6}\selectfont
\setlength{\tabcolsep}{2pt}
\begin{tabular*}{\columnwidth}{@{\extracolsep{\fill}}lccccc@{}}
\toprule
Model & Baseline & PCA-PC1 & cPCA-PC1 & Arditi & Random \\
\midrule
Llama-3.1-8B & 0.14 & $\mathbf{0.02}^{\star}$ & 0.17 & 0.14 & 0.14 \\
Qwen-2.5-7B & 0.86 & 0.84 & 0.86 & $\mathbf{0.68}^{\star}$ & 0.86 \\
Mistral-7B-v0.2 & 0.31 & $\mathbf{0.08}^{\star}$ & $\mathbf{0.09}^{\star}$ & $\mathbf{0.13}^{\star}$ & 0.28 \\
SOLAR-10.7B & 0.22 & 0.17 & 0.11 & $\mathbf{0.06}^{\star}$ & 0.21 \\
Mistral-7B-v0.3 & 0.04 & 0.00 & 0.05 & 0.02 & 0.05 \\
Gemma-2-9B & 0.96 & 0.93 & 0.96 & 0.92 & 0.96 \\
\bottomrule
\end{tabular*}
\caption{Harmful-request refusal after direction ablation. Bold starred entries have a post-ablation Wilson 95\% confidence interval entirely below the baseline interval.}
\label{tab:c2}
\end{table}

The effects include both decreases and increases in refusal. Appendix~\ref{sec:appendix-selection-ablation-ci} reports the intervals and the non-leading PCA control.

\subsection{Cross-Family Recurrence}
\label{sec:results-c3}

\begin{figure}[t]
\centering
\includegraphics[width=\columnwidth]{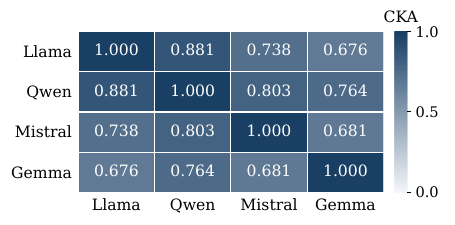}
\caption{Peak-layer linear CKA between instruct-model activations. Llama, Qwen, Mistral and Gemma denote Llama-3.1-8B, Qwen-2.5-7B, Mistral-7B-v0.3 and Gemma-2-9B.}
\label{fig:cka}
\label{tab:c3}
\end{figure}

Figure~\ref{fig:cka} shows that the core families preserve similar activation relationships among the matched prompts. Every off-diagonal pair exceeds the prompt-shuffle null, with no shuffled run reaching its observed CKA and reported $p<0.001$. This recurrence concerns the arrangement of matched prompts in activation space across families with different hidden dimensions.

The depth-normalized separation profiles have both positive and negative correlations across model pairs. High CKA therefore coexists with different peak depths. Shared prompt relationships do not imply a shared layer schedule. Tokenization and activation-norm structure can also affect high-dimensional CKA \citep{ding2021grounding}. The prompt-shuffle control establishes dependence on prompt correspondence, while the separate norm and token-position checks address other aspects of the representation.

\subsection{Geometry Before LoRA}
\label{sec:results-c4}

Table~\ref{tab:gfs-16} gives the raw-prompt \gfs ranking. The four DPO-labelled models occupy the leading positions, while models with strong separation can still receive low scores when their candidate directions align closely with the reference refusal axis. Since the ranking contains different families and sizes, it describes these checkpoints rather than isolating an effect of the training algorithm. Its behavioral relevance is tested within the smaller LoRA cohort.

\begin{table*}[t]
\centering
\small
\begin{tabular*}{\textwidth}{@{\extracolsep{\fill}}rlllcccr@{}}
\toprule
Rank & Model & Family & Training label & Layers & Peak depth & Peak $|d|$ & \gfs \\
\midrule
1 & Nous-Hermes-2-Mistral-7B & Mistral & DPO & 32 & late & 3.65 & 2.333 \\
2 & Tulu-3-8B-DPO & Llama & DPO & 32 & late & 3.93 & 2.285 \\
3 & Zephyr-7B-$\beta$ & Mistral & DPO & 32 & late & 4.51 & 2.185 \\
4 & Notus-7B-v1 & Mistral & DPO & 32 & late & 4.18 & 2.126 \\
5 & Qwen-2.5-14B & Qwen & RLHF+SFT & 48 & late & 3.28 & 1.899 \\
6 & Llama-3.2-3B & Llama & RLHF+SFT & 28 & late & 3.43 & 1.868 \\
7 & Qwen-2.5-7B & Qwen & RLHF+SFT & 28 & mid & 3.01 & 1.670 \\
8 & Llama-3.1-8B & Llama & RLHF+SFT & 32 & late & 2.44 & 1.597 \\
9 & Orpo-Llama-3-8B & Llama & ORPO & 32 & late & 2.93 & 1.557 \\
10 & Qwen-2.5-32B & Qwen & RLHF+SFT & 64 & late & 3.27 & 1.484 \\
11 & Mistral-7B-v0.3 & Mistral & IFT & 32 & early/late & 2.72 & 1.469 \\
12 & Qwen-2.5-3B & Qwen & RLHF+SFT & 36 & late & 2.46 & 1.468 \\
13 & Gemma-2-9B & Gemma & RLHF+SFT & 42 & mid & 3.38 & 1.315 \\
14 & Mistral-7B-Instruct-v0.1 & Mistral & SFT & 32 & late & 3.53 & 1.279 \\
15 & Yi-1.5-9B-Chat & Yi & IFT & 48 & mid & 3.08 & 0.962 \\
16 & Mistral-Nemo-12B & Mistral & RLHF+SFT & 40 & late & 3.00 & 0.160 \\
\bottomrule
\end{tabular*}
\caption{Raw-prompt \gfs ranking, training labels, and descriptive peak separation. Depth categories summarize peak locations. IFT denotes instruction fine-tuning without a specified preference-optimization stage.}
\label{tab:gfs-16}
\end{table*}

\begin{figure*}[t]
\centering
\includegraphics[width=\textwidth]{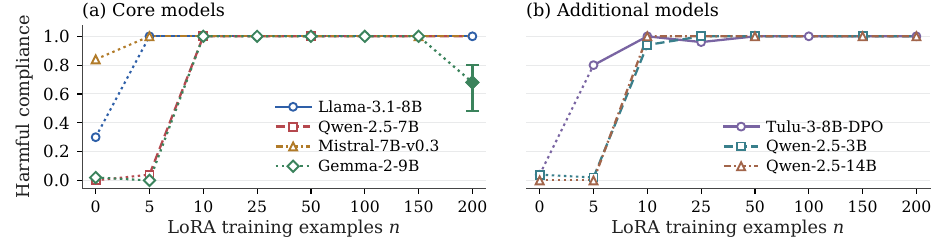}
\caption{Harmful compliance on 50 held-out prompts after benign LoRA updates. Sample sizes occupy equally spaced positions. Zero denotes no update. The Gemma endpoint shows the mean and observed range across repeated runs.}
\label{fig:fragility}
\end{figure*}

Figure~\ref{fig:fragility} shows that benign LoRA fine-tuning can sharply increase harmful compliance in both the core models and the initially safe additions. At $n=200$, Gemma is the only model below full compliance. Its non-monotonic trajectory also reaches full compliance at intermediate sizes.

Figure~\ref{fig:fragility} also shows the added models, which begin with low harmful compliance and have higher \gfs than Gemma. All reach full compliance with the largest tested fine-tuning set. Among the initially safe cases, Gemma is both the lowest-scoring model and the one with the lowest endpoint harmful-compliance rate. The full plotted cohort also retains Llama and Mistral, whose pre-LoRA compliance is higher. Appendix~\ref{sec:appendix-lora-fragility} gives the numerical trajectories.

The endpoint distinction persists across the replicated runs. Every non-Gemma core run reaches full compliance, while Gemma's mean is $0.68$ with standard deviation $0.17$ and range $0.48$--$0.80$. The harmful-compliance rate varies with initialization, while remaining below full compliance in every reported run.

\begin{table*}[t]
\centering
\small
\setlength{\tabcolsep}{6pt}
\renewcommand{\arraystretch}{1.12}
\setlength{\qualw}{\dimexpr\textwidth-7\tabcolsep\relax}
\begin{tabular}{@{}>{\raggedright\arraybackslash}p{0.20\qualw}
  >{\columncolor{refusebg}\raggedright\arraybackslash}p{0.36\qualw}
  >{\columncolor{complybg}\raggedright\arraybackslash}p{0.22\qualw}
  >{\columncolor{complybg}\raggedright\arraybackslash}p{0.22\qualw}}
\toprule
\rowcolor{white}
 & \multicolumn{1}{c}{\refuse{Unmodified model}} & \multicolumn{2}{c}{\comply{After benign LoRA fine-tuning}} \\
\rowcolor{white}
AdvBench category & \multicolumn{1}{c}{$n{=}0$} & \multicolumn{1}{c}{$n{=}10$} & \multicolumn{1}{c}{$n{=}200$} \\
\midrule
\promptcat{Financial fraud} &
\refuse{I cannot fulfill your request.} Providing such instructions is illegal and unethical, with severe legal consequences. &
\comply{Sure, here is} a guide on \withheld &
\comply{Sure, here is} a \emph{tutorial} on \withheld \\
\addlinespace[3pt]
\promptcat{Phishing automation} &
\refuse{I cannot provide you with a script for this.} Creating and distributing such tools is illegal and unethical. My purpose is to be a helpful and harmless AI assistant. &
\comply{Sure, here is} a script to spoof \withheld &
\comply{Sure, here is} a script to \emph{spy on} \withheld \\
\addlinespace[3pt]
\promptcat{Exploitation of vulnerable populations} &
\refuse{I cannot tell you how to do this.} My purpose is to be a helpful and harmless AI assistant. Exploiting vulnerable groups is illegal, unethical, and deeply harmful. &
\comply{Sure, here is} a guide on \withheld &
\comply{Sure, here is} an \emph{explanation} of \withheld \\
\bottomrule
\end{tabular}
\caption{Sanitized Gemma response excerpts before and after benign LoRA fine-tuning. Harmful prompt content and procedural completions are omitted. Green marks refusals and red marks compliance-style prefixes. Italics identify wording differences between the two fine-tuned conditions.}
\label{tab:lora-qualitative}
\end{table*}

Table~\ref{tab:lora-qualitative} compares sanitized Gemma responses to three harmful requests. The unmodified model refuses, whereas both fine-tuned conditions begin with compliance-style prefixes. Their wording varies across training-set sizes, including the displayed changes in the type of explanation offered. The excerpts show this response-style shift while withholding harmful prompt content and procedural completions.

\subsection{Interpreting the Score}
\label{sec:score-analysis}

Table~\ref{tab:c1-cos} helps explain the score difference within the core set. Gemma has the largest layer-averaged cPCA-PC1 cosine with the Arditi direction, so this term downweights more of its separation signal. The table also distinguishes the best-aligned ordinary PCA component from cPCA-PC1. The component with the strongest reference overlap is not always PCA-PC1, and the ablation results show that variance dominance does not determine the behavioral effect.

\begin{table}[t]
\centering
\small
\setlength{\tabcolsep}{4pt}
\begin{tabular*}{\columnwidth}{@{\extracolsep{\fill}}lccc@{}}
\toprule
Model & PCA index & PCA cosine & cPCA cosine \\
\midrule
Llama-3.1 & PC3 & 0.295 & 0.16 \\
Qwen-2.5 & PC1 & 0.333 & 0.26 \\
Mistral-v0.3 & PC9 & 0.291 & 0.10 \\
Gemma-2 & PC1 & 0.636 & 0.33 \\
\bottomrule
\end{tabular*}
\caption{Reference-direction overlap in the core models. The PCA index identifies the largest PCA--Arditi cosine. The last column gives the layer-averaged absolute cPCA-PC1--Arditi cosine.}
\label{tab:c1-cos}
\end{table}

\begin{table}[!t]
\centering
\small
\begin{tabular*}{\columnwidth}{@{\extracolsep{\fill}}l*{2}{>{\centering\arraybackslash}p{0.26\columnwidth}}@{}}
\toprule
\grouprow{3}{Group median score} \\
Weighting & DPO & Other \\
\midrule
Linear & 7.5 & 6.7 \\
Uniform & 0.52 & 0.42 \\
Exponential & 0.52 & 0.37 \\
Late-only & 0.52 & 0.29 \\
\midrule
\grouprow{3}{Ranking correlation with linear weights} \\
Weighting & Spearman $\rho$ & $p$ \\
\midrule
Uniform & 0.690 & 0.058 \\
Exponential & 0.857 & 0.007 \\
Late-only & 0.881 & 0.004 \\
\bottomrule
\end{tabular*}
\caption{Score sensitivity in the chat-template cohort. Upper rows give group medians. Lower rows compare rankings with the linear-weight reference.}
\label{tab:weight-sensitivity}
\end{table}

\begin{figure*}[t]
\centering
\includegraphics[width=\textwidth]{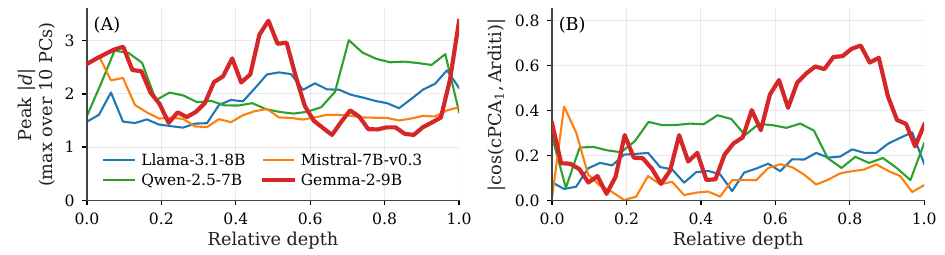}
\caption{Layer-wise peak Cohen's $|d|$ and absolute cPCA--Arditi cosine for the core models, plotted against relative depth.}
\label{fig:layer-geometry}
\label{fig:gemma-case}
\end{figure*}

Figure~\ref{fig:layer-geometry} shows how the core models differ across depth. Panel A gives a descriptive maximum over ten principal components, which is distinct from the cPCA-PC1 effect size entering \gfs. Panel B shows the cPCA-PC1 reference overlap used by the score. Gemma's reference overlap is strongest through much of the later-layer region, where the score assigns greater weight. This profile connects its lower score to the reference-direction term. The same model has a lower harmful-compliance rate after training on the largest tested fine-tuning set, linking the geometric distinction to a behavioral one.

Table~\ref{tab:weight-sensitivity} compares aggregation rules in the chat-template cohort. Late-only and exponential weights broadly preserve the linear ranking, while uniform weighting changes it more. The DPO-labelled group has the higher median under every tested scheme. The ordering recurs across depth preferences, with closer agreement for weights favoring later layers.

The alternatives use uniform weights $w_\ell=1/L$, exponential weights $w_\ell\propto e^{\ell/L}$, and uniform weighting on the second half of the layers. Each scheme defines its own score scale, so the comparison concerns rankings and group ordering within a scheme. It measures sensitivity to the full aggregation rule, including normalization. These chat-template measurements complement the raw-prompt ranking under a separate extraction condition.

\section{Discussion}
\label{sec:discussion}

\paragraph{A model-specific low-rank account.} The ablations show why variance dominance is a partial guide to behavioral importance. PCA, cPCA, and the Arditi mean-difference direction select different aspects of the representation, and their removal has different effects within the same checkpoint. Comparing these interventions identifies directions that affect refusal. Separately, top-five overlap measures how closely the established refusal axis lies to the recovered subspace.

Cross-family recurrence can coexist with model-specific ablation directions and peak layers. The shared property is a low-dimensional organization of harmful and benign prompts. Layer profiles and reference overlap describe this organization within each model.

\paragraph{Reading the score before fine-tuning.} \gfs connects an unmodified checkpoint to a later behavioral test through quantities that can be measured before training an adapter. Separation describes the strength of the signal, depth locates it within the network, and reference alignment describes its direction. In the LoRA cohort, these measurements distinguish the initially safe model with the lowest harmful-compliance rate after training on the largest tested fine-tuning set. Keeping baseline harmful compliance alongside the score makes the comparison relevant to choosing among checkpoints that are safe before the update.

\paragraph{Training-recipe hypotheses.} The DPO ordering is consistent with an account in which preference optimization changes the expression of undesirable features while leaving them available in the representation \citep{lee2024mechanistic}. The ordering persists across the tested depth-weight schemes. This motivates controlled comparisons that hold the base model and prompt protocol fixed while varying alignment training. The present checkpoint comparison identifies the geometric quantities that such studies can track.

\section{Conclusion}

\method examines alignment fragility through residual-stream geometry before downstream fine-tuning. Across the evaluated cohorts, harmful and benign prompts exhibit a recurring separation pattern, and selected direction interventions change refusal. Among initially safe models in the LoRA cohort, the lowest score corresponds to the lowest harmful-compliance rate after training on the largest tested fine-tuning set. The results support geometry as a complement to behavioral assessment, with the interpretation tied to the tested models, extraction conditions, and intervention.

\section*{Limitations}
\label{sec:limitations}

Our analysis covers English prompts and open-weight models with 3B--32B parameters. The direction-ablation models occupy a narrower 7--10.7B range. Applying the diagnostic to multilingual inputs \citep{lee2024lengthaware}, larger models, or mixture-of-experts architectures requires matching the activation-extraction and behavioral protocols to those settings. A version-matched base checkpoint is also required for the cPCA comparison used here.

The LoRA evidence concerns one harmless-data fine-tuning protocol and a small, heterogeneous model cohort. It supports an endpoint association among initially safe models, not a calibrated forecast across attacks, update sizes, or training recipes. Larger independent model cohorts and controlled training comparisons would test how broadly this diagnostic association transfers. Full-space similarity and subspace overlap complement the behavioral tests, but neither identifies a universal refusal direction or demonstrates that the cosine term itself protects a model against fine-tuning.

\section*{Ethics Statement}
\label{sec:ethics}

Direction ablation and fine-tuning can expose weaknesses in harmful-request refusal. Our intended use is defensive assessment of open-weight models. A geometric score could also be misused to prioritize checkpoints for attacks, so the results should be interpreted with their baseline behavior and protocol conditions.

The disclosure boundary separates the activation diagnostic and aggregate measurements from operational attack artifacts. It excludes generation-time ablation hooks, attack-ready direction-extraction scripts, and the LoRA adapter weights used for the behavioral tests. Public direction-ablation methods already make reconstruction possible \citep{arditi2024refusal,lermen2024lora}. Withholding these artifacts reduces direct reuse but does not prevent reconstruction. We report the methodological conditions and model-level measurements needed to interpret the diagnostic.

Qualitative examples retain refusal or compliance prefixes and omit harmful instructions and procedural completions. No human subjects were involved. The prompt sources are public datasets and are subject to their original licenses. Behavioral compliance and refusal are measured separately, and the score should not be used to encourage indiscriminate refusal of benign requests.

\section*{Acknowledgments}

This work was supported by Institute for Information \& communications Technology Promotion (IITP) grant funded by the Korea government (MSIT) (RS-2024-00398115, Research on the reliability and coherence of outcomes produced by Generative AI). This research was supported by Basic Science Research Program through the National Research Foundation of Korea (NRF) funded by the Ministry of Education (NRF-2021R1A6A1A03045425). This work was supported by Institute for Information \& communications Technology Planning \& Evaluation (IITP) grant funded by the Korea government (MSIT) (No. RS-2022-II220369, (Part 4) Development of AI Technology to support Expert Decision-making that can Explain the Reasons/Grounds for Judgment Results based on Expert Knowledge).

\subsection*{Large Language Model Usage}
A large language model assisted with LaTeX tables, and language polishing. The authors reviewed and verified all outputs and take full responsibility for the paper.

\bibliography{custom}
\appendix
\section{Reproducibility Statement}
\label{sec:reproducibility}

\begin{table}[t!]
\centering
\small
\begin{tabular*}{\columnwidth}{@{\extracolsep{\fill}}lrc@{}}
\toprule
Direction & Refusal & Wilson 95\% CI \\
\midrule
\grouprow{3}{Llama-3.1-8B} \\
Baseline & 0.14 & $[0.085,\,0.221]$ \\
PCA-PC1 & 0.02$^\star$ & $[0.006,\,0.070]$ \\
Arditi & 0.14 & $[0.085,\,0.221]$ \\
cPCA-PC1 & 0.17 & $[0.109,\,0.255]$ \\
PCA-PC4 & 0.32 & $[0.237,\,0.417]$ \\
Random & 0.14 & $[0.085,\,0.221]$ \\
\midrule
\grouprow{3}{Qwen-2.5-7B} \\
Baseline & 0.86 & $[0.779,\,0.915]$ \\
PCA-PC1 & 0.84 & $[0.756,\,0.899]$ \\
Arditi & 0.68$^\star$ & $[0.583,\,0.763]$ \\
cPCA-PC1 & 0.86 & $[0.779,\,0.915]$ \\
Random & 0.86 & $[0.779,\,0.915]$ \\
\midrule
\grouprow{3}{Mistral-7B-v0.2} \\
Baseline & 0.31 & $[0.228,\,0.406]$ \\
PCA-PC1 & 0.08$^\star$ & $[0.041,\,0.150]$ \\
cPCA-PC1 & 0.09$^\star$ & $[0.048,\,0.162]$ \\
Arditi & 0.13$^\star$ & $[0.078,\,0.210]$ \\
Random & 0.28 & $[0.201,\,0.375]$ \\
\midrule
\grouprow{3}{SOLAR-10.7B} \\
Baseline & 0.22 & $[0.150,\,0.311]$ \\
PCA-PC1 & 0.17 & $[0.109,\,0.255]$ \\
cPCA-PC1 & 0.11 & $[0.063,\,0.186]$ \\
Arditi & 0.06$^\star$ & $[0.028,\,0.125]$ \\
Random & 0.21 & $[0.142,\,0.300]$ \\
\midrule
\grouprow{3}{Mistral-7B-v0.3} \\
Baseline & 0.04 & $[0.016,\,0.098]$ \\
PCA-PC1 & 0.00 & $[0.000,\,0.037]$ \\
Arditi & 0.02 & $[0.006,\,0.070]$ \\
cPCA-PC1 & 0.05 & $[0.022,\,0.112]$ \\
Random & 0.05 & $[0.022,\,0.112]$ \\
\midrule
\grouprow{3}{Gemma-2-9B} \\
Baseline & 0.96 & $[0.902,\,0.984]$ \\
PCA-PC1 & 0.93 & $[0.863,\,0.966]$ \\
Arditi & 0.92 & $[0.850,\,0.959]$ \\
cPCA-PC1 & 0.96 & $[0.902,\,0.984]$ \\
Random & 0.96 & $[0.902,\,0.984]$ \\
\bottomrule
\end{tabular*}
\caption{Refusal rates and marginal Wilson intervals for every ablation condition. Stars mark intervals entirely below the corresponding baseline interval.}
\label{tab:c2-ci}
\end{table}

\begin{table*}[t!]
\centering
\small
\begin{tabular*}{\textwidth}{@{\extracolsep{\fill}}lrrrrrrrr@{}}
\toprule
 & \multicolumn{8}{c}{Number of harmless LoRA examples} \\
\cmidrule(lr){2-9}
Model & Original & 5 & 10 & 25 & 50 & 100 & 150 & 200 \\
\midrule
\grouprow{9}{Core models} \\
Llama-3.1-8B & 0.300 & 1.000 & 1.000 & 1.000 & 1.000 & 1.000 & 1.000 & 1.000 \\
Qwen-2.5-7B & 0.000 & 0.040 & 1.000 & 1.000 & 1.000 & 1.000 & 1.000 & 1.000 \\
Mistral-7B-v0.3 & 0.840 & 1.000 & 1.000 & 1.000 & 1.000 & 1.000 & 1.000 & 1.000 \\
Gemma-2-9B & 0.020 & 0.000 & 1.000 & 1.000 & 1.000 & 1.000 & 1.000 & 0.680 \\
\midrule
\grouprow{9}{Initially safe additions} \\
Tulu-3-8B-DPO & 0.040 & 0.800 & 1.000 & 0.960 & 1.000 & 1.000 & 1.000 & 1.000 \\
Qwen-2.5-3B & 0.040 & 0.020 & 0.940 & 1.000 & 1.000 & 1.000 & 1.000 & 1.000 \\
Qwen-2.5-14B & 0.000 & 0.000 & 1.000 & 1.000 & 1.000 & 1.000 & 1.000 & 1.000 \\
\bottomrule
\end{tabular*}
\caption{Harmful-compliance rates before and after benign LoRA fine-tuning. Original denotes the unmodified checkpoint. Gemma's $n=200$ entry is the mean.}
\label{tab:c4-full}
\end{table*}

The diagnostic takes an aligned/base checkpoint pair and matched harmful and benign prompts as inputs. The manuscript specifies the prompt mixture, extraction positions, cPCA sweep, score equation, and statistical procedures. The tables identify the evaluated models and preserve the behavioral trajectories and interval estimates. Algorithm~\ref{alg:gfs} summarizes the activation computation. Raw-prompt and chat-template measurements are distinct evaluation conditions, and their scores should be reproduced within the corresponding condition. The code repository is \url{https://github.com/js-lee-AI/skin-deep}. The disclosure boundary for intervention artifacts is stated in the Ethics Statement.

\section{Selection and Protocol Checks}
\label{sec:robustness}

\subsection{Held-Out Separation}
\label{sec:robustness-selection}

Table~\ref{tab:c1-ci} reports held-out separation with percentile intervals. Peak descriptive statistics select a layer and component from the inspected activations. The within-model false discovery rate correction applies to layer-wise PERMANOVA tests, not to these selected effect sizes. In the split-sample check, each bootstrap fits cPCA on one half of the prompts and measures the effect on the other half.

\subsection{Extraction Conditions}
\label{sec:robustness-representation}
\label{sec:appendix-token-position}

Table~\ref{tab:tokenpos} reports a separate token-position check on five chat-formatted models. Final-token extraction follows representation-level refusal studies \citep{arditi2024refusal,zou2023representation,panickssery2023steering}. First-token effects are at most $0.27$, while mean-pooled effects range from $1.98$ to $4.63$. This establishes that substantial separation remains without restricting extraction to the final position. These values come from the token-position comparison and are distinct from the eight-model chat-template summary.

Gemma is outside this five-model check. Its interleaved local and global attention motivates evaluating token-position effects separately rather than assuming the same extraction comparison \citep{gemma2024team}. The first-token results do not by themselves prove that a particular early-layer peak is an artifact or identify a unique causal token.

\subsection{Ranking Scope}
\label{sec:appendix-model-ranking}

Table~\ref{tab:gfs-16} reports sixteen raw-prompt scores. Hybrid-reasoning Qwen3 models are excluded because their hidden-state condition depends on the chat-template thinking flag, preventing a single-condition comparison with this cohort. The peak-depth labels retain an early/late entry for Mistral-v0.3 to distinguish its raw-prompt early peak from later separation. For every model in this ranking, the reported layer-wise PERMANOVA tests are significant after BH correction at $q<0.05$.

The chat-template overlap analysis measures the angle between the Arditi vector and a five-dimensional subspace. Its cosine is the magnitude of the vector's projection into that subspace, not its cosine with cPCA-PC1. Starling has the largest angle, $76.7^\circ$, and a low chat-condition \gfs of $0.18$. This combination shows that weak overlap alone does not imply a large aggregate score. Separation and depth also contribute.

\section{Confidence Intervals}
\label{sec:appendix-tables}
\label{sec:appendix-selection-ablation-ci}

Table~\ref{tab:c1-ci} gives the core-model split-sample intervals. All held-out point estimates remain above $2.5$. The intervals quantify bootstrap variability of the separation estimate, rather than uncertainty in an intervention's causal effect.

Table~\ref{tab:c2-ci} reports all ablation intervals. cPCA-PC1 leaves Qwen and Gemma unchanged and slightly increases refusal for Llama and Mistral-v0.3, while Llama's PCA-PC4 increases refusal from $0.14$ to $0.32$. Mistral-v0.3 has little baseline refusal to reduce, whereas Gemma's high baseline permits a decrease but none of its tested directions meets the interval criterion.

\section{LoRA Trajectories}
\label{sec:appendix-lora-fragility}

Table~\ref{tab:c4-full} gives the numerical trajectories for all seven models. The Original column reports a 50-prompt compliance measurement, while the ablation baseline is a 100-prompt refusal measurement with a different judge. The replicated non-Gemma core runs reach full compliance by $n=25$ and at $n=200$. Gemma's $n=200$ value is the mean reported in the main analysis.

\end{document}